\documentclass[10pt,journal]{IEEEtran}
\usepackage[utf8]{inputenc}
\usepackage{stfloats}
\usepackage{float}
\usepackage{amsmath,amsfonts,amssymb}
\usepackage{algorithm}
\usepackage{algorithmic}
\usepackage{graphicx}
\usepackage{booktabs}
\usepackage[nocompress]{cite}
\usepackage{color}
\usepackage{hyperref}
\usepackage{url}
\usepackage{amsthm}
\theoremstyle{plain}
\newtheorem{theorem}{Theorem}
\newtheorem{lemma}{Lemma}
\newtheorem{corollary}{Corollary}
\newtheorem{proposition}{Proposition}
\newtheorem{assumption}{Assumption}
\newtheorem{costmodel}{Model}
\theoremstyle{remark}\newtheorem{remark}{Remark}

\begin{document}

\title{H-Mem$+$: Source-Augmented Hierarchical Memory for Budgeted Evidence Matching}

\author{Shaoning~Zeng, Wen~Huang, Haoran~Sun, Jianhang~Zhou, Bob~Zhang, and~Yong~Xu%
\thanks{S. Zeng and W. Huang contributed equally to this work and share first authorship.}%
}

\IEEEtitleabstractindextext{%
\begin{abstract}
Long-term dialogue question answering requires matching a query to source records that may be distributed across sessions. Finding the highest-scoring record can still leave out evidence needed for the answer. H-Mem+ uses source-augmented hierarchical memory to combine candidates from episode summaries and original dialogue windows. It reranks them with a local cross-encoder, and packs the linked source turns within a fixed evidence budget of 2,048 BERT wordpieces. We derive conditional bounds on ancestor-index retention and hierarchical routing scores, and distinguish missing candidates from evidence omitted during ranking and packing. On LoCoMo10, H-Mem+ answers all 1,986 questions, achieving factual F1 of 0.5995 on 1,540 non-adversarial questions and 75.11\% correct refusal on 446 adversarial questions. On a retrospective paired sample of 50 questions from one development conversation, H-Mem+ uses 95.04\% fewer retrieval-plus-answer API tokens than our HiGMem integration, with factual F1 of 0.4695 versus 0.3911 on the 40 factual questions in the sample. The token comparison covers online retrieval and answering and excludes memory construction and local computation. Source code of this paper can be found at https://github.com/UESTC-HZ/H-Mem.

\end{abstract}

}

\maketitle
\IEEEdisplaynontitleabstractindextext
\IEEEpeerreviewmaketitle

\section{Introduction}

\begin{figure*}[!t]
\centering
\includegraphics[width=\textwidth]{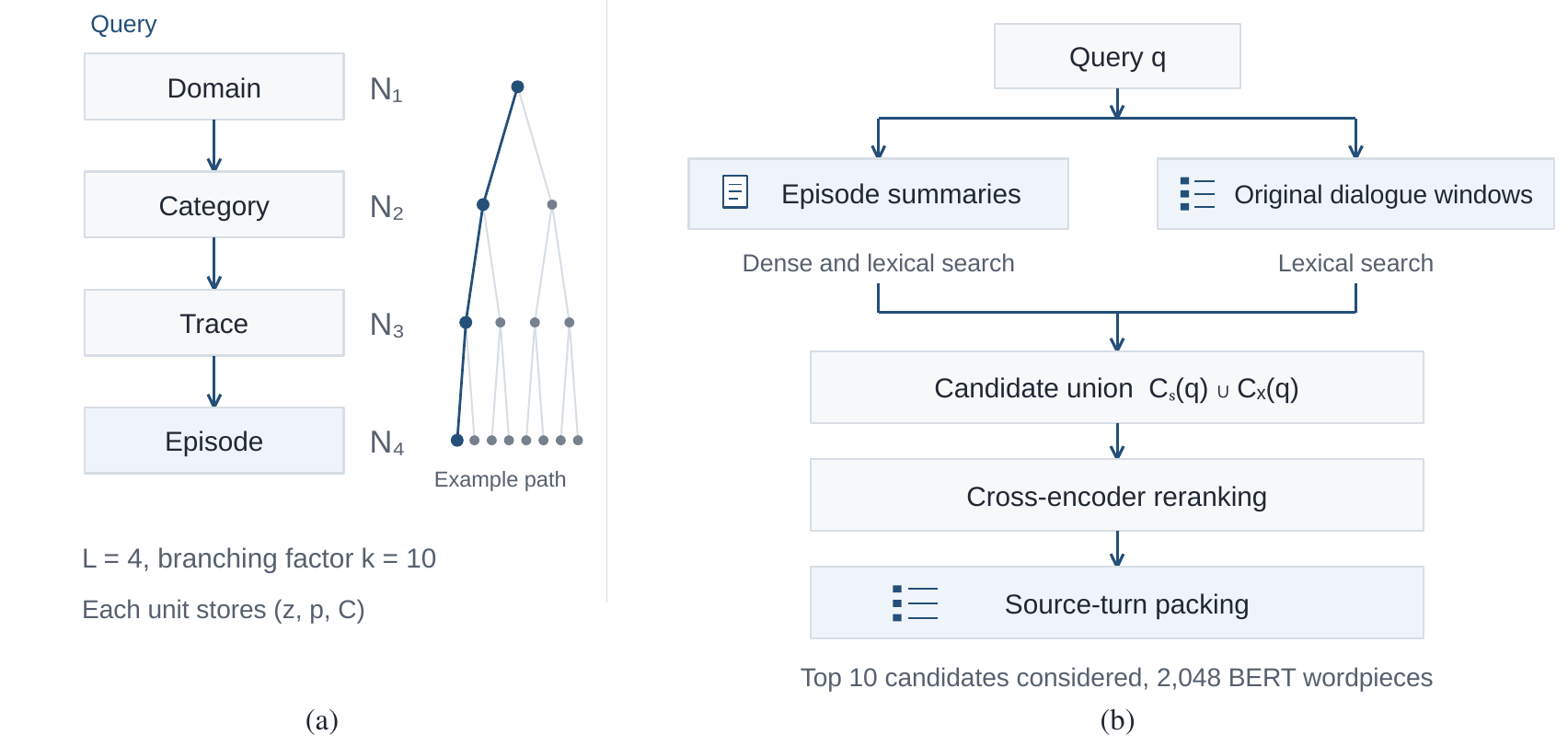}
\caption{Hierarchical index and evidence selection. (a) \emph{H-Mem index and routing.} The inherited index organizes units at four levels. A unit stores a semantic vector $z$, a position index $p$, and child indices $C$. The highlighted path illustrates coarse-to-fine routing through a subset of branches. (b) \emph{H-Mem+ evidence selection.} H-Mem+ combines candidates from episode summaries and original dialogue windows, reranks the union with a cross encoder, and packs linked source turns. Packing considers the ten highest-ranked candidates within a 2,048-wordpiece budget (see Section~\ref{sec:pipeline}).}
\label{fig:hmem_architecture}
\end{figure*}

\IEEEPARstart{L}{ong-term} dialogue memory helps large language model (LLM) agents maintain coherent interactions across sessions \cite{wei2025planner3d}. An assistant can recall preferences stated in an earlier conversation, carry forward decisions about an ongoing task, and connect a new request with relevant past events \cite{zhang2025memorysurvey,park2023generativeagents}. Such continuity is valuable when a task evolves over weeks or months, as new information can be interpreted in the context of the user's earlier goals. Longitudinal benchmarks such as LoCoMo evaluate these capabilities through questions and summaries that draw on events across sessions \cite{maharana2024locomo}.
Reliable access to earlier exchanges is therefore critical factor to building assistants that support sustained interactions.

Retrieval-based memory systems organize interaction histories for later access. MemoryBank retrieves stored memories by similarity, while RAPTOR and HiGMem use summary hierarchies to guide access to text \cite{zhong2024memorybank,sarthi2024raptor,cao2026higmem}. Information bottleneck and multi-resolution analysis provide complementary views of what a compact representation can retain at different scales \cite{tishby2000ib,mallat1989multiresolution}. Coarse-to-fine search uses those scales to narrow the candidate set and limit scoring work, as in H-Mem's position-index routing \cite{fleuret2001coarsetofine,sun2026h}. LoCoMo and LongMemEval assess long-term memory on questions drawn from interactions across sessions \cite{maharana2024locomo,wu2024longmemeval}. However, retrieving the highest-scoring record does not show that the selected set covers the evidence needed to answer. Identifying who attended an event and when it occurred can require records from different sessions or subtrees. Candidate generation can miss a relevant record entirely. Ranking and packing can also omit a record already retrieved. The question is how retrieval can recover the records that jointly support an answer under a fixed context budget. 
Addressing this question requires distinguishing candidate availability from evidence delivery, and evaluating answer quality separately from retrieval-score guarantees and resource consumption.

We propose H-Mem+, a source-augmented hierarchical memory for budgeted evidence matching in long-term dialogues. H-Mem narrows the search through top-down index routing \cite{sun2026h}. H-Mem+ retains the summary hierarchy and adds a separate retrieval path over original dialogue windows, so source evidence can enter the candidate pool even when summary search misses it. A local cross-encoder reranks the combined candidates, and the packing step admits deduplicated source turns within a fixed evidence budget. The contributions are as follows.
\begin{itemize}
\item \textbf{Conditional Hierarchical Memory Guarantees.} We derive conditional results for ancestor-index retention, routing scores,
depth, and averaging stability. We show why preserving the best similarity score can leave set recall low.
\item \textbf{Source-Augmented Evidence Selection.} We develop a source-augmented retrieval pipeline and distinguish evidence
missing from the candidate set from evidence omitted during ranking and packing.
\item \textbf{Answer Quality and API-Token Efficiency.} We evaluate answer quality on LoCoMo10 and API token use on fixed paired
samples. Matched ablations assess the source candidate path and the reranker.
\end{itemize}

The remainder of this paper is organized as follows. Related work and the gap are stated in Section~\ref{sec:rw-memory}. Section~\ref{sec:guarantees} presents the architecture and derives the four results, each architectural component followed by the analysis of that component. Section~\ref{sec:experiments} reports the measured results, the model-level validation, and a discussion of what the evidence does and does not establish. Section~\ref{sec:conclusion} concludes this paper.

\section{Related Work}

\subsection{Long-Term Memory for LLM Agents}
\label{sec:rw-memory}

Retrieval-augmented generation established the pattern of storing external text and fetching it at
inference time \cite{lewis2020rag,karpukhin2020dpr}, and agent memory inherits it while adding write and update, because memory is produced by the agent's own interaction stream and changes as that stream
continues \cite{zhang2025memorysurvey}. Memory is accordingly treated as one of the core modules that must keep adapting if an agent is to learn over its lifetime \cite{zheng2026lifelong}, and memory-augmented agents have been built for embodied open-world tasks as well as for dialogue \cite{wang2025jarvis1}.

The conference baselines also include the LoCoMo full-context setting \cite{maharana2024locomo} and ReadAgent \cite{lee2024readagent}, which combines gist memory with access to the original passages.
Work on agent memory can be grouped by the structure imposed on the store. \emph{Flat} systems keep all units at one level and retrieve by similarity. MemoryBank encodes memories as vectors and retrieves by
cosine similarity \cite{zhong2024memorybank}, Generative Agents scores a single stream by recency,
importance and relevance \cite{park2023generativeagents}, and Mem0 combines vector and graph storage
\cite{chhikara2025mem0}. Their cost is structural. The similarity computation is against the whole store, so cost grows with the number of stored units, which is the regime quantified in
Section~\ref{sec:efficiency}. \emph{Structured} systems trade that cost for organization, relating
units through explicit edges as in A-MEM's self-organizing note network \cite{xu2025amem} or
HippoRAG's hippocampal-index analogue \cite{gu2024hipporag}. They represent relations faithfully but
must maintain and traverse the structure, and their retrieval guarantees are not characterized.
\emph{Hierarchical} systems are the closest to this paper. MemGPT pages information between a small
main context and external storage under an operating-system analogy \cite{packer2023memgpt}. MemoryOS develops that analogy into explicit tiers \cite{kang2025memoryos}, and RAPTOR recursively clusters and summarizes text into a tree, retrieving at several levels of abstraction
\cite{sarthi2024raptor}. Work in this direction continues. HiMem builds a two-level
episode--note hierarchy for long-horizon dialogue and reconsolidates knowledge that later turns
contradict \cite{zhang2026himem}, and HiGMem uses event summaries as semantic anchors so that the
model reads only the turns worth reading \cite{cao2026higmem}. Both report LoCoMo results with
GPT-4o-mini as their main backbone, HiGMem adding a Qwen2.5-3B comparison, whereas our protocol
(see Section~\ref{sec:setup}) runs every method under one setting across six open-weight backbones, so
cross-paper headline numbers are not directly comparable. A third system has to be separated by name.
H-Mem \cite{ye2026hmem} appeared in the same EACL volume as our conference version, shares our
acronym, evaluates on the same benchmark and cites this line of work, but is a distinct system;
throughout this paper H-MEM means ours \cite{sun2026h}. It identifies coarse evidence localization as a limitation of hierarchical retrieval.
Section~\ref{sec:retrieval} characterizes this limitation through a bound on the gap to exhaustive
search and conditions for exact descent. HiGMem's motivation bears directly on
this paper. It observes that hierarchical systems which rank solely by vector similarity return
redundant evidence sets, in which many superficially similar turns add little recall while lowering
precision and raising answer-stage cost. That is the qualitative form of the problem whose
quantitative form we give in Section~\ref{sec:retrieval}. What these systems share with H-MEM is the
multi-level organization. What they lack is an analysis saying how much information a level may discard, how good a descending retrieval is
relative to exhaustive search, or how deep the hierarchy should be.

\subsection{Information-Theoretic and Multi-Scale Views}
\label{sec:rw-it}

The information bottleneck formalizes representation as a trade-off between compressing an input and
retaining what is relevant to a target \cite{tishby2000ib}, both as an explanatory account of deep
network training \cite{tishby2015deep} and as a training principle through variational bounds
\cite{alemi2016vib}, with the data processing inequality constraining information along any
deterministic chain \cite{cover2005elements} and caveats arising when the deterministic case is pushed too far
\cite{kolchinsky2018caveats,amjad2020ib}. Amjad and Geiger \cite{amjad2020ib}
identify ill-posed or piecewise-constant IB objectives in deterministic networks and limitations
of using that objective to characterize desirable representations. We therefore distinguish
conditional index-entropy statements from claims about task-optimal semantic representations.
Hu et al. \cite{hu2024ibsurvey} survey IB theory and applications to pattern analysis and representation
learning. Task-specific extensions include the two-stream bottleneck of Wu and Deng
\cite{wu2024tib} for unknown-object detection, where standard and reverse objectives separate
relevant and irrelevant representations. That visual-learning setting differs from our memory
pipeline, but illustrates why an information objective must be tied to its downstream task. Representing a signal at several resolutions is equally classical. Scale-space theory builds a family of smoothed representations
\cite{witkin1987scalespace} and multiresolution analysis formalizes the resulting decomposition
\cite{mallat1989multiresolution}, with hierarchical feature pyramids as the learned counterpart
\cite{lin2017fpn} and agglomerative clustering as the algorithmic one \cite{mullner2011modern}.

This literature supplies our tools, but its objects are not ours. An information bottleneck analysis
often studies encoders against a task variable. Our focus instead is a \emph{stack} of coarsening layers
whose unit counts are fixed by a branching factor and whose entries are reached by index rather than by
exhaustive scoring. Conversely, the memory work above invokes compression informally without stating a
bound. What is missing is the composition of the two, namely a chain of deterministic abstractions whose retention, total compression and information ordering can be written in terms of the architecture's
own quantities. Section~\ref{sec:flow} also makes explicit that the data processing inequality
constrains the \emph{direction} of information flow only, and therefore cannot by itself support an
optimality claim.

The closest work to this paper is a recent formalism that decomposes hierarchical memory into
extraction, coarsening and traversal operators and uses that decomposition to compare eleven existing
systems \cite{talebirad2026theory}. Its notation is compatible with ours, and it supplies a shared
vocabulary for the design space. What it leaves open is the quantitative side, namely how much a level
retains, how much worse a descending retrieval is than exhaustive search, and how deep a hierarchy
should be; its authors also record the static-hierarchy assumption as the central limitation of the
account. The same openness is acknowledged from a systems perspective, one recent lifecycle-memory
study stating that its design does not depend on a particular number of layers being uniquely
optimal \cite{han2026retention}.

\subsection{Coarse-to-Fine Retrieval and the Cost of Memory}
\label{sec:rw-scale}

\begin{figure}[!htbp]
\centering
\includegraphics[width=\columnwidth]{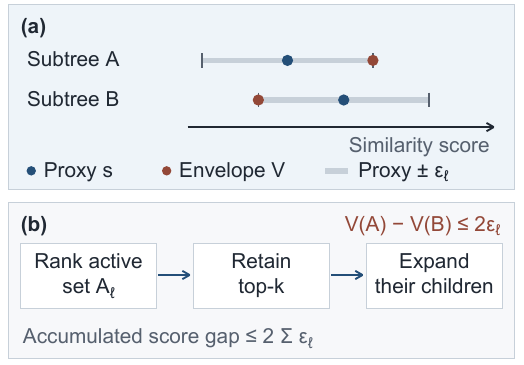}
\caption{Subtree-envelope error for one routing transition. (a) \emph{Proxy scores and subtree envelopes.} In the active set $A_\ell$, subtree $A$ has the largest envelope $V$ and subtree $B$ has the highest proxy score $s$. The horizontal intervals show the admissible error $\varepsilon_\ell$. (b) \emph{One routing transition.} The proxy ranking and the two envelope errors bound the score gap by $2\varepsilon_\ell$. Routing retains the top-$k$ candidates and expands their children to form the next active set. Summing the transition bounds gives Theorem~\ref{th:retrieval}.}
\label{fig:envelope}
\end{figure}

Retrieval from a multi-level structure is coarse-to-fine search. Descend through progressively finer partitions instead of scanning the finest level, an idea with a long history in detection
\cite{fleuret2001coarsetofine} and with approximate nearest-neighbour search as its scalable form,
from locality-sensitive hashing \cite{indyk1998ann} through graph and quantization indexes
\cite{malkov2018hnsw} to large-scale vector search libraries
\cite{johnson2019billion,douze2025faiss}. These methods bound distance error to the true nearest
neighbour under probabilistic assumptions about the index. Our Assumption~\ref{as:envelope} is a
different condition in the same spirit, in that it bounds how well a coarse unit's similarity proxies the best similarity in its subtree, but its purpose is to make the bound \emph{estimable on a query
sample}, so that the retrieval guarantee can be checked rather than assumed. A further difference is
that our units are semantic abstractions produced by a language model rather than geometric partitions
of a metric space, which is why the condition is stated as a measurable error rather than as a
probabilistic guarantee. Figure~\ref{fig:envelope} illustrates one such routing transition.

Memory decay and reinforcement have likewise been studied extensively. The forgetting curve gives an exponential account of retention over time \cite{ebbinghaus1913memory} and is adopted directly by MemoryBank. Consolidation describes how initially labile traces stabilize, and catastrophic forgetting motivates regularization- and rehearsal-based update rules \cite{kirkpatrick2017ewc}, with
complementary learning systems explaining why an architecture would separate a fast, detail-rich store
from a slow, structured one \cite{mcclelland1995complementary} and hierarchical process memory
proposing temporally receptive hierarchies as the organizing principle \cite{hasson2015hierarchical}.
A recent preprint takes the complementary route on the update side, fusing a sliding window with an
exponential moving average for preference weights and justifying the fused estimator statistically, once
as a posterior expectation and once as a simplified Kalman filter \cite{sun2025pamu}. That justification
concerns the estimator under a noise model, whereas the multiplicative feedback rule analysed below is
not an estimator, so the two are not substitutes. Our Theorem~\ref{th:stability} isolates one mechanism by which a hierarchy obtains stability for free, since averaging over children damps a sparse update, which is a property of the aggregation operator
rather than of a learning rule, and which is what separates content stability from the structural
observation that a coarser layer has fewer units to touch in the first place.

\noindent\label{sec:rw-rerank}Retrieval quality is itself a two-stage problem in the broader literature,
and that literature bears directly on what this paper measured. A first stage generates candidates with
a cheap representation -- sparse, as in BM25 \cite{robertson2009bm25}, or dense, as in dual-encoder
retrieval \cite{karpukhin2020dpr} -- and a second scores the candidates against the query with a model
that reads both. Re-ranking with a cross encoder \cite{nogueira2019passage} is the standard form of the
second stage, and the measured gap between the two stages is what motivates it. A bi-encoder compresses
the query and the passage separately, so it can only compare where the compressed representations happen
to align.

For long-term dialogue, candidate generation must balance broad semantic matching
with the recovery of specific names, dates, and relations. H-Mem+ searches episode
summaries and original dialogue windows, then jointly scores the question and each
candidate before packing its linked source turns. The ablations in
Section~\ref{sec:final-ablations} test the source candidate path and the choice of
reranker. Separately, Proposition~\ref{pr:setrecall} shows how a routing rule that
commits to one subtree can preserve the best match while losing other relevant
items. This explains why a score certificate alone leaves the evidence set
undetermined.

\subsection{Benchmarks and Positioning}
\label{sec:rw-bench}

Evaluation has driven much of the progress. LoCoMo \cite{maharana2024locomo} provides multi-session
dialogues with question--answer pairs spanning recall, multi-hop synthesis, temporal reasoning,
open-domain knowledge and unanswerable queries, and is the benchmark we use. LongMemEval \cite{wu2024longmemeval} targets long-term interactive memory in assistants. A parallel line measures
whether long inputs are used at all. RULER \cite{hsieh2024ruler} probes effective context length and the ``lost in the middle'' effect \cite{liu2024lostinthemiddle} shows that a long context can be present yet
unused, which is a reason to evaluate memory architectures on retrieval rather than on context length.
None of these benchmarks measures the quantities our analysis predicts, such as per-layer entropy and
retention, subtree envelope error, the dominance profile of content over structural change, and the
information ordering across layers, so benchmark scores cannot stand in for evidence about the theory.

The gap this paper addresses is thus not the absence of hierarchical memory systems, of which there are
several, but the absence of an analysis of them. We do three things differently. First, we state the
retention, retrieval, depth and stability results in terms of the architecture's own parameters, so
that design decisions follow from a cost model rather than from tuning. Second, we state each
assumption as a quantity that can be estimated on data, such as layer occupancy, subtree envelope error and minority change, to support empirical assessment of the bounds. Third, we separate
measured from model-level evidence throughout, because performance on a benchmark does not by itself
establish why a system works.

\section{Guarantees for Hierarchical Memory}
\label{sec:guarantees}

\subsection{Overall Architecture and Notation}

The inherited H-Mem index organizes interaction memory into $L$ layers of semantic resolution
(see Section~\ref{sec:rw-it} for the multi-scale lineage this follows). We formalize it as $\mathcal{H}=(\mathcal{L},\mathcal{A},\mathcal{E})$,
where $\mathcal{L}=\{\mathcal{M}_1,\dots,\mathcal{M}_L\}$ is the layer sequence,
$\mathcal{A}=\{A_1,\dots,A_{L-1}\}$ are the abstraction (aggregation) operators with
$A_\ell:\mathcal{M}_{\ell+1}\to\mathcal{M}_\ell$, and $\mathcal{E}=\{E_\ell\}$ are the
per-layer encoders.

\begin{figure}[!htbp]
\centering
\includegraphics[width=\columnwidth]{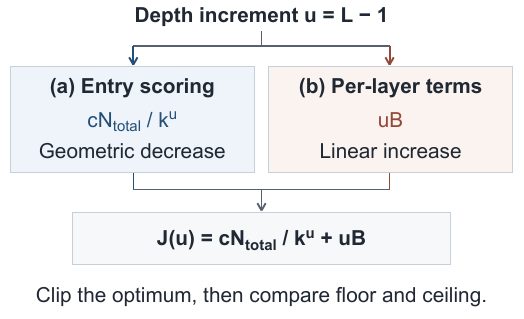}
\caption{Depth trade-off in the routing surrogate (see Eq.~\eqref{eq:objective}). As $u=L-1$ increases, (a) the entry scoring cost decays geometrically and (b) the per-level cost and abstraction penalty grow linearly. Their sum $J(u)$ is strictly convex. The feasible integer solution follows by clipping the stationary point to the feasible interval and comparing its floor and ceiling.}
\label{fig:depth}
\end{figure}

Layer $\ell$ contains $\mathcal{M}_\ell=\{m_{\ell,1},\dots,m_{\ell,N_\ell}\}$ memory units.
Throughout, we index layers so that \textbf{$\ell=1$ is the coarsest and $\ell=L$ the finest},
i.e.\ $N_1<N_2<\cdots<N_L$, and the parent--child direction runs from coarse to fine. Each unit
$m_{\ell,i}$ stores a semantic vector $z_{\ell,i}=E_\ell(\cdot)\in\mathbb{R}^{D}$, a positional
index $p_{\ell,i}$, and child indices $C_{\ell,i}$ pointing to its sub-units in layer $\ell+1$.
With branching $k_\ell$ we have $N_{\ell+1}=N_\ell k_\ell$, and for uniform branching $k$,
\begin{equation}
N_\ell \;=\; N_{total}\,k^{\,\ell-L},
\label{eq:layer_size}
\end{equation}
where $N_{total}=N_L$ is the number of stored episodes. Equation~\eqref{eq:layer_size} is an idealized balanced-tree model, not an invariant of the measured stores. H-Mem+ additionally retains source turns and their provenance links outside this summary index. Figure~\ref{fig:depth} previews the routing surrogate that combines this layer-size model with a penalty for abstraction (see Section~\ref{sec:depth}).

H-Mem \cite{sun2026h} uses four semantic scales, from global domains and thematic categories
to traces that summarize local sequences and episodes that store fine-grained interactions
and the user profile. The top three layers index the episode content. Semantic aggregation
defines the scales, and a shared sentence encoder supports similarity comparisons across them.
In the balanced model, $N_\ell$ counts the units at each scale and $k$ gives the ratio between
adjacent scales (see Eq.~\eqref{eq:layer_size}).

\subsection{Hierarchical Encoding and Information Compression}

\begin{figure}[!htbp]
\centering
\includegraphics[width=\columnwidth]{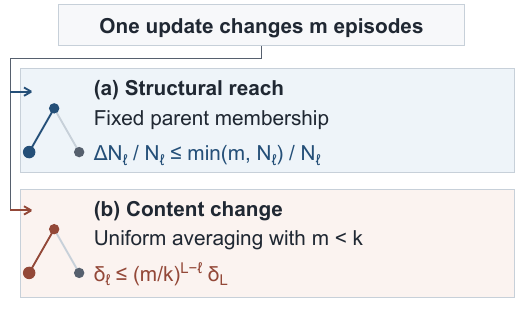}
\caption{Structural reach and content change after an update to $m$ episodes. (a) With fixed parent membership, each changed leaf can touch one ancestor per level, giving an upper bound of $\min(m,N_\ell)/N_\ell$ on the touched fraction. (b) Under uniform averaging and minority change, the maximum scalar weight change contracts toward coarser levels by the factor $m/k$ at each transition. The two branches show analytical bounds under their respective assumptions.}
\label{fig:stability}
\end{figure}

The implementation extracts layer-specific text and embeds those texts; a parent can
be reused by a similarity-based insertion rule. It does not recompute every parent embedding
as the mean of its children. We therefore distinguish the implemented summary index from the
following conditional aggregation model. In that model,
\begin{equation}
\begin{aligned}
z_{\ell,i}&=E_\ell\!\left(\frac{1}{k_{\ell,i}}\sum_{j\in C_{\ell,i}}z_{\ell+1,j}\right),\\
w_{\ell,i}&=\frac{1}{k_{\ell,i}}\sum_{j\in C_{\ell,i}}w_{\ell+1,j}.
\end{aligned}
\label{eq:aggregation}
\end{equation}
Determinism suffices for a data-processing statement; averaging is an additional assumption
needed for the stability result. Neither condition is established for free-form LLM summary
updates merely by organizing nodes into a tree. Figure~\ref{fig:stability} illustrates the distinction between touching an ancestor and changing its scalar weight (see Section~\ref{sec:update}).

To make the unit-count entropy statements precise, fix a store and a distribution over its
episodes. Let $U_L$ be a sampled episode identifier and $U_\ell$ its unique ancestor identifier
at layer $\ell$. For the retention analysis only, $\mathcal{M}_\ell$ in an entropy or mutual
information expression denotes $U_\ell$, not the full vector-valued contents of that layer.
There are at most $N_\ell$ possible identifiers and $U_\ell$ is a deterministic function of
$U_{\ell+1}$. Thus these results quantify loss of index resolution. They do not bound the
semantic information in a source span or the number of API tokens it takes to encode it.
Assume positive denominator entropies whenever a retention ratio is used.

For adjacent layers, define retention as the mutual information normalized by the
entropy of the finer identifier:
\begin{equation}
R_\ell \;=\; \frac{I(\mathcal{M}_\ell;\mathcal{M}_{\ell+1})}{H(\mathcal{M}_{\ell+1})}.
\label{eq:retention}
\end{equation}

\begin{lemma}[Deterministic aggregation]
\label{lem:det}
$R_\ell = H(\mathcal{M}_\ell)/H(\mathcal{M}_{\ell+1}) \le 1$, with equality if and only if
$A_\ell$ is injective on the support of $\mathcal{M}_{\ell+1}$.
\end{lemma}
\begin{proof}
Determinism gives $I(\mathcal{M}_\ell;\mathcal{M}_{\ell+1})=H(\mathcal{M}_\ell)$.
By the entropy chain rule, equality holds exactly when the finer identifier is
recoverable from its parent on the support.
\end{proof}

\begin{assumption}[Layer occupancy]
\label{as:uniform}
For an exact count-based bound at a given transition, the finer ancestor identifier is uniformly distributed: $H(U_{\ell+1})=\log N_{\ell+1}$. Near-uniform occupancy alone does not imply this equality; otherwise the actual entropy must remain in the denominator.
\end{assumption}

\begin{theorem}[Information retention bound]
\label{th:retention}
Under Assumption~\ref{as:uniform} and branching $k_\ell$,
\begin{equation}
R_\ell \;\le\; \frac{\log N_\ell}{\log N_{\ell+1}}
\;<\; 1\qquad (k_\ell>1).
\label{eq:retention_bound}
\end{equation}
\end{theorem}
\begin{proof}
Apply the entropy ratio in Lemma~\ref{lem:det}, using $H(U_\ell)\le\log N_\ell$,
$H(U_{\ell+1})=\log N_{\ell+1}$, and $N_{\ell+1}=k_\ell N_\ell$.
\end{proof}

\begin{corollary}[Telescoping retention]
\label{co:telescope}
$\displaystyle R_{total}=\prod_{\ell=1}^{L-1}R_\ell=\frac{H(\mathcal{M}_1)}{H(\mathcal{M}_L)}$. Under Assumption~\ref{as:uniform},
$R_{total}\le \log N_1/\log N_{total}$.
\end{corollary}
\begin{proof}
Multiply the adjacent entropy ratios and use $H(U_1)\le\log N_1$ and
$H(U_L)=\log N_{total}$.
\end{proof}

\begin{corollary}[Depth and index-resolution loss]
\label{co:depthcomp}
For a balanced tree with uniform branching $k>1$, a uniform finest identifier, and
$N_{total}>1$,
\begin{equation}
1-R_{total}(L)\ \ge\ \frac{(L-1)\log k}{\log N_{total}}.
\label{eq:loss}
\end{equation}
Equality requires uniform occupancy also at the coarsest layer.
\end{corollary}
\begin{proof}
Substitute $\log N_1=\log N_{total}-(L-1)\log k$ in the bound of
Corollary~\ref{co:telescope}. Balanced branching and uniform leaves give uniform
coarse occupancy and attain equality.
\end{proof}

\noindent\phantomsection\label{sec:flow}With fixed parents, the interaction record $X$ and its sampled
episode identifier give the Markov chain $X\to U_L\to\cdots\to U_1$. Data processing yields
$I(U_1;X)\le\cdots\le I(U_L;X)$, with $I(U_\ell;X)\le H(U_\ell)\le\log N_\ell$.
These inequalities quantify index resolution. Extending them to summary contents requires
a Markov chain among those summaries, a condition that separate prompts from $X$ do not
establish. H-Mem+'s source path retains $X$, subject to candidate and packing omissions
(see Proposition~\ref{pr:source-bypass} and Eq.~\eqref{eq:packing-loss}). Task-specific
rate--distortion or information-bottleneck optimality remains open.

\subsection{Top-Down Retrieval and Approximation Optimality}
\label{sec:retrieval}

Retrieval descends the hierarchy. At layer $\ell$ the algorithm scores candidate units against
the query $q$, keeps the best, and continues only inside the selected unit's children, so that
similarity is never evaluated against all $N_{total}$ episodes.

\begin{algorithm}[t]
\caption{Conditional hierarchical routing (H-Mem reference)}
\label{alg:retrieval}
\begin{algorithmic}[1]
\REQUIRE query $q$, layers $\mathcal{M}_1..\mathcal{M}_L$, beam width $k\ge1$
\STATE $A_1\leftarrow\mathcal{M}_1$
\FOR{$\ell=1$ \TO $L-1$}
\STATE $B_\ell\leftarrow\operatorname{top}\text{-}k(A_\ell,q)$
\STATE $A_{\ell+1}\leftarrow\bigcup_{m\in B_\ell}C_m$
\ENDFOR
\RETURN $\arg\max_{m\in A_L}\operatorname{sim}(q,m)$
\end{algorithmic}
\end{algorithm}

To bound the quality of this descent we need a condition stating that a coarse unit's similarity
to the query is a usable proxy for the best similarity reachable anywhere in its subtree. For a
unit $m\in\mathcal{M}_\ell$, define its \emph{subtree envelope}
\begin{equation}
\begin{aligned}
V_\ell(m) &= \max_{c\,\in\,\mathrm{leaves}(\mathrm{desc}(m))}\mathrm{sim}(q,c),\\[2pt]
\Phi &= \max_{m\in\mathcal{M}_L}\mathrm{sim}(q,m).
\end{aligned}
\label{eq:envelope}
\end{equation}

\begin{assumption}[Envelope approximation]
\label{as:envelope}
There exist $\varepsilon_\ell\ge 0$ such that for every query $q$ and every $m\in\mathcal{M}_\ell$,
$|\mathrm{sim}(q,m)-V_\ell(m)|\le\varepsilon_\ell$. Note $\varepsilon_L=0$, since at the episode
layer the envelope is the similarity itself.
\end{assumption}

The envelope condition measures how closely the learned proxy estimates the best score
in an active subtree (as shown in Fig.~\ref{fig:envelope}). Its error can be estimated
on sampled queries.

\begin{theorem}[Retrieval approximation]
\label{th:retrieval}
Under Assumption~\ref{as:envelope}, the episode $m^\ast$ returned by
Algorithm~\ref{alg:retrieval} satisfies
\begin{equation}
\mathrm{sim}(q,m^\ast) \;\ge\; \Phi - 2\sum_{\ell=1}^{L-1}\varepsilon_\ell .
\label{eq:retrieval_bound}
\end{equation}
\end{theorem}
\begin{proof}
Let $M_\ell=\max_{m\in A_\ell}V_\ell(m)$. Initially $M_1=\Phi$.
Choose $a\in A_\ell$ attaining $M_\ell$ and let $b$ maximize proxy similarity over
$A_\ell$; $b$ is retained because $k\ge1$. The envelope condition gives
$V_\ell(b)\ge\operatorname{sim}(q,b)-\varepsilon_\ell
\ge\operatorname{sim}(q,a)-\varepsilon_\ell\ge M_\ell-2\varepsilon_\ell$.
Its children are expanded, so $M_{\ell+1}\ge M_\ell-2\varepsilon_\ell$.
At the leaves $V_L=\operatorname{sim}$; induction proves the bound.
\end{proof}
\begin{corollary}[Exact routing under exact envelopes]
\label{co:exact}
If every envelope error is zero, Algorithm~\ref{alg:retrieval} returns a maximum-similarity
leaf. This does not imply exact answers to multi-source questions.
\end{corollary}
\begin{remark}[Reader-independent conditional statements]
\label{rm:reader_free}
The routing and coverage statements contain no reader parameters, but their assumptions
and the constructed memory still depend on the encoders and extraction process.
Reader-independent notation is not a guarantee of backbone-invariant F1.
\end{remark}
The bound applies to any nonempty beam and controls only the best similarity score
returned by the reference router. It gives no guarantee on beam nesting, set recall,
or H-Mem+'s packed evidence and answer accuracy.

\begin{proposition}[Set recall is not implied by the score bound]
\label{pr:setrecall}
Let the stored units have similarities $s_1\ge s_2\ge\cdots$ and let $T_k$ be the $k$ units of
highest similarity. Let a descending retrieval commit at layer $2$ to a subtree $S$ and return
$\mathcal{D}$, the $k$ highest-scoring units inside $S$. Suppose the members of $T_k$ are spread over
$b$ distinct subtrees at that layer. Then $\Delta=\Phi-\max_{m\in\mathcal{D}}\operatorname{sim}(q,m)=0$
whenever the top unit lies in $S$, while $|\mathcal{D}\cap T_k|\le\lceil k/b\rceil$ when the members
are evenly spread, and $|\mathcal{D}\cap T_k|=1$ when each of them lies in a different subtree.
\end{proposition}

\begin{proof}
$\Delta$ depends only on $s_1$ and on whether $1\in S$; the overlap depends on where the remaining
$k-1$ units lie. Placing unit $1$ in $S$ and units $2,\dots,k$ outside it gives $\Delta=0$ and
$|\mathcal{D}\cap T_k|=1$, so no bound on $\Delta$ implies a bound on the overlap.
\end{proof}

The consequence is limited but useful: the score inequality alone places no bound on
set recall. Additional assumptions about how the required evidence is distributed among
candidate subtrees are necessary.

A second consequence follows once a question needs more than one memory. Proposition~\ref{pr:setrecall}
concerns a single answer-bearing unit; a multi-hop or enumeration question needs several, and each one
must survive retrieval on its own.

\begin{proposition}[Cardinality-$c$ coverage]
\label{pr:cardinality}
Let a question require $c$ distinct memories and let $p_i$ be the probability that memory $i$ reaches
the reader. Then the probability that all $c$ are available is at most $\min_i p_i$, and equals
$\prod_i p_i$ when the retrieval events are independent. Under that independence condition and a common per-item recall $p$, joint coverage is $p^c$. Without independence this expression is not an upper bound.
\end{proposition}

\begin{proof}
The joint event $\bigcap_i\{i\text{ retrieved}\}$ is contained in each of its marginals, which gives
the bound; independence factors the joint probability into the product.
\end{proof}

Under independent per-item retrieval with common recall $p$, reaching joint coverage
$1-\delta$ requires $p\ge(1-\delta)^{1/c}$. This is a conditional requirement, not a
prescription for a fixed candidate depth: the mapping from candidate count to evidence
coverage has to be measured, and packing can discard retrieved sources.

\begin{corollary}[Retrieval cost]
\label{co:cost}
Scoring cost is $N_1+\sum_{\ell=2}^{L} k_\ell$ for single-path descent and
$N_1+\sum_{\ell=2}^{L} k\,k_\ell$ when expanding top-$k$ per layer, against $N_L$ for exhaustive
search. In the historical H-Mem configuration \cite{sun2026h} ($a$ domains, $100$ children per unit
below the top layer, top-$k$ with $k=10$), exhaustive search costs $O(a\cdot10^6\cdot D)$ while
hierarchical retrieval costs $O((a+k\cdot300)\cdot D)$.
\end{corollary}

\subsection{H-Mem+: Source-Augmented Evidence Selection}
\label{sec:pipeline}
H-Mem+ augments the hierarchical summary index with a separate candidate path over
original dialogue windows. It retains each text turn with its session date, speaker, and
source identifier. Each stored episode links to its contiguous source turns. Imported
episodes are admitted only when their source sequence belongs to the current dialogue.
The source path retrieves candidates directly from the original turns and bypasses
summary compression and top-down routing.

\noindent\label{sec:pipeline-candidates}Let $C_s(q)$ be the union of the dense top-50 and
idf-overlap top-50 episode summaries. Each candidate is rendered with its linked source
turns. Independently, form a window around each original turn using its immediate previous
and next turns within the same session, and retain the lexical top-50 windows, $C_x(q)$.
The candidate documents are $C(q)=C_s(q)\cup C_x(q)$, with source provenance retained.
Ties use deterministic input order. There are at most 150 candidate documents before
source deduplication. The implementation uses BERT-based embeddings for summary search
and a local \texttt{ms-marco-MiniLM-L-6-v2} cross encoder for question--document scoring.

\noindent\label{sec:pipeline-rerank}The cross encoder orders the candidates jointly with the
question. Only the ten highest-scoring documents are considered for packing. In that order,
the algorithm proposes their not-yet-selected source turns, sorts the resulting source set
chronologically, and admits the proposal only if its rendered context has at most 2,048
local BERT wordpieces. A rejected document does not cause the algorithm to scan candidates
ranked below ten. The reader sees deduplicated original turns and their provenance, not a
concatenation of every candidate summary. This is a deterministic feasible packing rule,
not an optimizer of a submodular objective or a knapsack approximation scheme.

\begin{proposition}[Source bypass and candidate coverage]
\label{pr:source-bypass}
For a finite-valued interaction record $X$, summary $S$, and target answer $Y$,
$H(Y\mid X,S)\le H(Y\mid S)$. If $S=f(X)$ is deterministic, then
$I(X;(X,S))=H(X)$. Further, if $\Gamma(C)$ denotes the source identifiers linked to a
candidate set, then $\Gamma(C_s)\subseteq\Gamma(C_s\cup C_x)$.
\end{proposition}
\begin{proof}
Conditioning cannot increase conditional entropy; the joint representation includes $X$
explicitly. The final inclusion is the set-union identity.
\end{proof}
These are availability statements. They neither say that summaries alone preserve all facts
nor that the larger candidate pool improves the packed context: added documents can displace
useful candidates in the top ten, and a useful source can exceed the packing budget.

\begin{proposition}[Candidate and packing omissions]
\label{pr:packing-omission}
For a query with a nonempty set $G(q)$ of required source identifiers, let
$A(q)=\Gamma(C(q))$ and let $E(q)\subseteq A(q)$ be the packed source identifiers. Define
$r_C=|G\cap A|/|G|$ and $r_E=|G\cap E|/|G|$. Then
\begin{equation}
1-r_E=\frac{|G\setminus A|}{|G|}+\frac{|G\cap(A\setminus E)|}{|G|}.
\label{eq:packing-loss}
\end{equation}
If $a_i=\Pr(i\notin A)$ and $b_i=\Pr(i\in A,\ i\notin E)$ for each $i\in G$, then
$\Pr(G\subseteq E)\ge\max\{0,1-\sum_{i\in G}(a_i+b_i)\}$.
\end{proposition}
\begin{proof}
Because $E\subseteq A$, the missing evidence partitions disjointly into evidence absent
from the candidates and evidence present but not delivered. The probability inequality is
the union bound over per-source omissions and requires no independence assumption.
\end{proof}
The reference evidence is used only to define a diagnostic, never as retrieval input.
The second term includes both top-ten ranking and budget rejection. Thus reducing summary
loss or improving candidate recall alone does not guarantee higher F1. The experiment
reports final answers rather than claiming this decomposition was measured on every query.

\noindent\label{sec:pipeline-answer}Local retrieval and reranking make no generative API calls.
For a query $q$, the metered online cost is
\begin{equation}
T_{\rm online}(q)=\sum_{c\in\mathcal{C}_{\rm ret}(q)\cup\mathcal{C}_{\rm ans}(q)}
(T^{\rm in}_c+T^{\rm out}_c),
\label{eq:online-tokens}
\end{equation}
where calls include failed attempts with known usage. For H-Mem+, $\mathcal{C}_{\rm ret}$ is
empty; memory-writing cost is a separate quantity, not zero. The BERT-wordpiece packing
constraint is not an API-token bound because the tokenizers differ. Nor does it bound the
reader's output length or total prompt, which also contains instructions and the question. 

\subsection{Depth of a Conditional Routing Surrogate}
\label{sec:depth}

In the balanced occupancy model, deeper hierarchies lose more index resolution (see Corollary~\ref{co:depthcomp}) and are cheaper to enter
(see Eq.~\eqref{eq:layer_size}), but they add scoring work at every level and discard information.
We make this trade-off explicit.

\begin{costmodel}[Retrieval cost]
\label{mo:cost}
Scoring costs $c$ per similarity computation. The entry layer is scored once in full. At each subsequent level the algorithm scores not the whole layer but the children of the $k$ retained
candidates, up to $k\,k_\ell$ units. The full-beam scoring surrogate is
\begin{equation}
\begin{aligned}
C(L) &= c\bigl(N_1+\sum_{\ell=1}^{L-1}k\,k_\ell\bigr)\\[2pt]
&\overset{k_\ell=k}{=} c\bigl(N_{total}k^{-(L-1)}+(L-1)k^2\bigr).
\end{aligned}
\label{eq:cost}
\end{equation}
Actual scoring can be smaller when fewer nodes are available; this expression is the idealized full-beam cost. The distinction matters. Scanning whole layers would give $\sum_\ell N_\ell$, which grows with depth,
whereas scoring candidate sets gives Eq.~\eqref{eq:cost}, whose first term decays geometrically. It is
this decay, not the per-level overhead, that makes a deep hierarchy worth building.
\end{costmodel}

The two terms therefore pull in opposite directions, the first decaying geometrically with depth and the second growing
linearly, which is what allows an interior optimum to exist at all (see Fig.~\ref{fig:depth}).

Combining with Corollary~\ref{co:depthcomp}, we define a routing surrogate using the linear index-loss expression, with penalty $\lambda\ge0$. This surrogate is not H-Mem+'s measured API-token objective and is not an exact semantic-loss objective outside the uniform balanced model. Writing $u=L-1$,
\begin{equation}
J(u) \;=\; \underbrace{c\,N_{total}k^{-u}}_{\text{entry cost}}
\;+\; u\underbrace{\Bigl(ck^2+\frac{\lambda\log k}{\log N_{total}}\Bigr)}_{\textstyle B}.
\label{eq:objective}
\end{equation}

\begin{theorem}[Unique optimal depth]
\label{th:depth}
For $c>0$, $N_{total}>1$, and $k>1$, $J(u)$ is strictly convex. Its stationary point on the real line is
\begin{equation}
L^\ast \;=\; 1+\log_k\!\left(\frac{c\,N_{total}\ln k}{\,ck^2+\lambda\,\dfrac{\log k}{\log N_{total}}\,}\right).
\label{eq:optimal_depth}
\end{equation}
\end{theorem}
\begin{proof}
$k^{-u}=e^{-u\ln k}$ is strictly convex and adding the linear term $Bu$ preserves strict convexity, so
the first-order condition is necessary and sufficient. Solving it gives Eq.~\eqref{eq:optimal_depth}.
If the stationary point is negative, the constrained optimum on $u\ge0$ is zero. The bounded feasible solution is specified below.
\end{proof}

The depth is additionally bounded by feasibility, since $N_1=N_{total}k^{-(L-1)}\ge1$ requires $L\le L_{\max}=1+\lfloor\log_k N_{total}\rfloor$. Clip the stationary depth to $[1,L_{\max}]$ for the continuous optimum. For integer depth, evaluate $J$ at the feasible floor and ceiling of that clipped value and choose the lower objective; ordinary rounding is not in general exact.

\begin{corollary}[Depth grows with memory size]
\label{co:depthsize}
$L^\ast$ increases monotonically (logarithmically) in $N_{total}$. Larger memories justify deeper hierarchies.
\end{corollary}

\begin{corollary}[Illustrative four-layer working point]
\label{co:four}
With $k=10$, $N_{total}=10^6$ and $c=1$, Eq.~\eqref{eq:optimal_depth} gives $L^\ast=5.36$ at
$\lambda=0$, $L^\ast=4.94$ at $\lambda=10^3$, and $L^\ast=4.12$ at $\lambda=10^4$. In this surrogate only,
$L=4$ corresponds to the regime in which avoiding abstraction loss dominates
retrieval cost, i.e.\ $\lambda\approx1.2\times10^4$ in this cost model, which does not identify the deployment penalty from data or prove the measured four-layer store optimal.
\end{corollary}

The complementary question, how wide each level should be, has been answered at fixed depth. With cost
proportional to $Lkb$ and $n=b^L$, the cost falls as $b/\ln b$, is minimised at $b=e$, and the depth
then follows as $\lceil\ln n/\ln b^\ast\rceil$ \cite{talebirad2026theory}. Depth and branching are
separate design variables, so the two analyses do not compete, and they differ in what they can say to
a store that grows. That account fixes depth only indirectly, through the branching optimum, and its
prescription for branching is independent of $n$. The objective above is instead optimised over depth,
is strictly convex with a unique minimiser, and moves with the store, prescribing a deeper hierarchy as
$N_{total}$ grows. Our cost model also carries the branching factor in both of its terms, since
$B=ck^2+\lambda\log k/\log N_{total}$, so it implies an optimum for $k$ as well, namely the solution of
$2ck+\lambda/(k\log N_{total})=cN_{total}k^{-L}$. That optimum depends on $N_{total}$, which a
prescription of $b=e$ for every store cannot reproduce, and the difference traces to the geometric
entry term that their cost model does not carry.

\subsection{Memory Update and Hierarchy Stability}
\label{sec:update}

In the conditional averaging model, new episodes are inserted at layer $L$. Higher layers are refreshed by
re-applying Eq.~\eqref{eq:aggregation} along the affected paths. Memory weights follow
\begin{equation}
\begin{aligned}
w_i(t) &= w_i(t-1)\,f(\text{feedback}),\\[2pt]
f &= \begin{cases}
>1, & \text{user approves (reinforcement)},\\
e^{-\lambda\Delta t}, & \text{no feedback (Ebbinghaus decay)},\\
<1, & \text{user rebuts (weakening)}.
\end{cases}
\end{aligned}
\label{eq:update}
\end{equation}
The mechanism adjusts external memory weights only. The base language model is never updated.

Stability has two notions that behave in opposite directions across the hierarchy, so it is
important to separate \emph{structural} stability, which counts how many units an update touches,
from \emph{content} stability, which measures how far each unit's representation moves
(see Fig.~\ref{fig:stability}).

\begin{proposition}[Structural change]
\label{pr:struct}
For a fixed tree with unchanged parent memberships, if one update event modifies $m$ episodes at layer $L$, the number of touched units at layer $\ell$
is at most $\min(m,N_\ell)$.
\end{proposition}
\begin{proof}
Each modified episode has exactly one ancestor at layer $\ell$, so the touched set is a subset of
those $m$ ancestors and is also bounded by $N_\ell$.
\end{proof}

Because $N_\ell$ increases with $\ell$, the \emph{relative} share $\Delta N_\ell/N_\ell$ is in fact
larger at coarse layers. Structural counting therefore cannot support the claim that coarse layers are more stable. That claim has to be made about content, which is where we turn next.

\begin{assumption}[Minority change]
\label{as:minority}
In a single update event, at most $m$ children of any unit change, with $m<k$, and each changed
child moves by at most $\delta_L$.
\end{assumption}

\begin{theorem}[Content stability increases towards coarse layers]
\label{th:stability}
Under the uniform-weight aggregation of Eq.~\eqref{eq:aggregation} and
Assumption~\ref{as:minority}, the per-unit scalar weight change at layer $\ell$ obeys
\begin{equation}
\delta_\ell \;=\; \max_i\|\Delta w_{\ell,i}\| \;\le\; \Bigl(\frac{m}{k}\Bigr)^{L-\ell}\delta_L .
\label{eq:stability}
\end{equation}
\end{theorem}
\begin{proof}
At most $m$ children of any unit change (Assumption~\ref{as:minority}), and the aggregate moves by
their average, so $\|\Delta w_{\ell,i}\|\le(m/k)\,\delta_{\ell+1}$. Iterating from $\ell=L-1$ down to $1$ gives Eq.~\eqref{eq:stability}.
\end{proof}

Since $m<k$ the damping compounds with depth. The domain layer's perturbation is smaller than the
episode layer's by the factor $(k/m)^{L-1}$, which is the sense in which abstract layers are the stable
ones.

\begin{remark}[Update certificate and its limits]
\label{co:retrieval_stability}
The averaging result bounds scalar weights. The same argument applies to representation
vectors only with fixed child membership and nonexpansive linear aggregation; an encoder
with operator norm $K_\ell$ introduces an additional factor $K_\ell$ at that transition.
Free-form summary rewrites need not satisfy either model. After an update,
Theorem~\ref{th:retrieval} can be reapplied if the new envelope errors are bounded, yielding
a new certificate $2\sum_\ell\varepsilon'_\ell$. A change in that certificate is not a
bound on the difference between two selected answers. Such a claim would additionally
require control of leaf-score drift, proxy Lipschitz constants, and ranking margins.
\end{remark}


\section{Experimental Evaluation}
\label{sec:experiments}
We evaluate H-Mem+ as the new method of this paper. The primary evidence is a complete
LoCoMo10 run and a separate paired token substudy. Historical H-Mem results, earlier
configuration diagnostics, and conditional-model calculations are kept distinct.

\subsection{Experimental Setup}
\label{sec:setup}
LoCoMo10 \cite{maharana2024locomo} contains ten conversations and 1,986 questions:
841 single-hop, 282 multi-hop, 321 temporal, 96 open-domain, and 446 adversarial.
We average normalized token-overlap F1 and BLEU-1 over the 1,540 factual questions;
BLEU-1 is unigram precision without a brevity penalty. Adversarial items have no factual
answer span, so we report the proportion recognized as correct refusals separately.
A question-weighted mean is the arithmetic mean of per-question scores, while a dialogue
macro mean gives equal weight to each conversation. Neither is the unweighted mean of
four category scores.

The final H-Mem+ run uses DeepSeek-V4.1-Flash-1017 with temperature zero, a 4,096-token completion cap, and local seed zero. No API seed is supplied. Local retrieval
uses \texttt{bert-base-uncased} embeddings and
\texttt{cross-encoder/ms-marco-MiniLM-L-6-v2} on CPU. The exact evidence-selection rule is
specified in Section~\ref{sec:pipeline}. The 2,048-wordpiece context cap is distinct from the
API completion cap. Experiments use text turns with session-date provenance, without adding
image captions unavailable to the other integrations. Dates are rendered explicitly to
resolve expressions such as yesterday, and readers are instructed to give short answers.

Conv-26 was used during method development. Its 199 successful responses are reused unchanged.
Parameters were then frozen; a fixed 90-question check across the remaining nine conversations
preceded completing their remaining questions. Existing summaries and embeddings were reused,
with source sequences filtered to their own dialogue. Historical caches contained some episodes
from other conversations; their removal is a provenance correction and prevents attributing all
old-to-new differences to a single algorithmic component. Each API response and usage record is
saved durably; interrupted execution resumes without reanswering successful questions.

\subsection{Full-Benchmark Accuracy}
\label{sec:main}
\begin{table*}[!t]
\centering
\footnotesize
\setlength{\tabcolsep}{5pt}
\caption{Ten-dialogue LoCoMo results with DeepSeek-V4.1-Flash-1017.}
\label{tab:current-main}
\begin{tabular}{lrrrrrrrr}
\toprule
System & Questions & Single-hop & Multi-hop & Temporal & Open-domain & Factual F1 & BLEU-1 & Correct refusal \\
\midrule
LoCoMo~\cite{maharana2024locomo} & 1,986 & 0.4791 & 0.3185 & 0.2339 & 0.1717 & 0.3794 & 0.3020 & 357/446 \\
ReadAgent~\cite{lee2024readagent} & 1,986 & 0.4618 & 0.3166 & 0.3076 & 0.2508 & 0.3899 & 0.3229 & 216/446 \\
MemoryBank~\cite{zhong2024memorybank} & 1,986 & 0.3645 & 0.2562 & 0.1751 & 0.1215 & 0.2900 & 0.2380 & 351/446 \\
A-Mem~\cite{xu2025amem} & 1,986 & 0.3118 & 0.2103 & 0.2562 & 0.1646 & 0.2725 & 0.3049 & 288/446 \\
H-Mem~\cite{sun2026h} & 1,986 & 0.1207 & 0.0803 & 0.0749 & 0.0909 & 0.1019 & 0.0757 & 396/446 \\
MemFit~\cite{piehl2026memfit} & 1,986 & 0.4509 & 0.2175 & 0.3782 & 0.1726 & 0.3757 & 0.4150 & 248/446 \\
HiGMem~\cite{cao2026higmem} & 1,986 & 0.5417 & 0.4479 & 0.5777 & 0.1792 & 0.5094 & 0.5533 & 328/446 \\
\textbf{H-Mem+} & \textbf{1,986} & \textbf{0.6887} & \textbf{0.4348} & \textbf{0.5877} & \textbf{0.3418} & \textbf{0.5995} & \textbf{0.6387} & \textbf{335/446} \\
\bottomrule
\end{tabular}
\end{table*}

Table~\ref{tab:current-main} uses the same ten dialogues and 1,986 question IDs for every system.
Its overall F1 and BLEU-1 average the 1,540 factual questions; correct refusal uses the separate
446-question adversarial denominator. Open-domain cells cover all 96 questions in that category,
even when another category is pending.
For H-Mem+, all 1,986 expected unique questions succeed. H-Mem+ reaches factual F1 $0.5995$ and
BLEU-1 $0.6387$; dialogue-macro factual F1 is $0.5993$. It correctly refuses $335/446$
adversarial questions ($75.11\%$), with no empty evidence contexts. Factual F1 on the nine
non-development conversations is $0.6036$, compared with $0.5628$ on conv-26. This split
supports applicability beyond the development conversation but is not an independent-dataset
or multi-seed evaluation. Single-hop F1 is $0.6887$, followed by temporal F1 $0.5877$, multi-hop F1 $0.4348$,
and open-domain F1 $0.3418$ (see Table~\ref{tab:current-main}). The gaps from single-hop
are $0.2539$ for multi-hop and $0.3469$ for open-domain questions, identifying these
categories as the principal remaining accuracy weaknesses.
We compare H-Mem+ with the saved ten-dialogue rerun of the H-Mem conference method
(see Table~\ref{tab:current-main}). The saved H-Mem rerun obtains factual F1 $0.1019$
and BLEU-1 $0.0757$, compared with $0.5995$ and $0.6387$ for H-Mem+; however, it correctly
refuses more adversarial questions ($396/446$, or $88.79\%$, versus $335/446$, or $75.11\%$).
The factual-accuracy improvement therefore coexists with a 13.68-percentage-point decrease
in correct refusal. The different historical construction and reader settings
prevent attributing the change to a single algorithmic component. The completed LoCoMo full-context integration contributes all 1,986 question outcomes,
including four metered length-limit failures scored as zero. Its factual F1 is $0.3794$,
BLEU-1 is $0.3020$, and correct refusal is $357/446$. The completed A-Mem and MemFit
integrations are also reported in Table~\ref{tab:current-main}. ReadAgent, MemoryBank,
and HiGMem each report outcomes for all ten dialogues. Completion counts every
retained question outcome, including failed answers scored as zero.

\begin{figure*}[!t]
\centering
\includegraphics[width=0.85\textwidth]{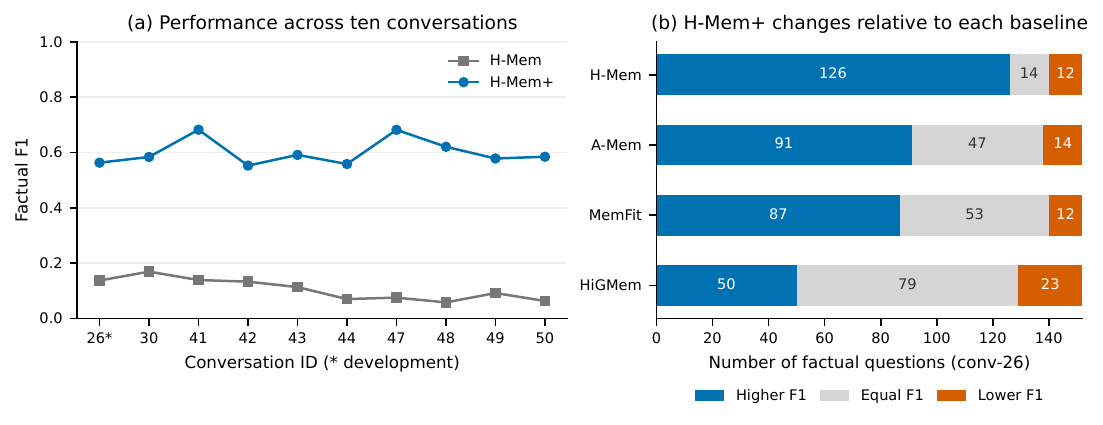}
\caption{Measured accuracy beyond aggregate scores. Left: factual F1 on each of ten conversations,
with the development conversation marked by an asterisk. Right: the numbers of conv-26 factual
questions for which H-Mem+ has higher, equal, or lower F1 than each baseline. The paired view
exposes regressions hidden by a mean improvement. Historical integration and failure-recovery
differences remain as stated in the text.}
\label{fig:empirical-accuracy}
\end{figure*}
To assess whether the aggregate improvement is concentrated in a few conversations, we
compare factual F1 separately across all ten conversations. H-Mem+ ranges from $0.5522$
on conv-42 to $0.6820$ on conv-41, whereas the saved H-Mem rerun ranges from $0.0576$
on conv-48 to $0.1688$ on conv-30. H-Mem+ has higher factual F1 on every conversation;
the absolute gains range from $0.4146$ on conv-30 to $0.6066$ on conv-47. Thus the
higher aggregate score is accompanied by gains across all evaluated conversations,
although the magnitude varies substantially (see Figure~\ref{fig:empirical-accuracy}(a)).

We further compare individual answers on the same 152 factual questions in conv-26.
Relative to H-Mem, H-Mem+ improves F1 on 126 questions, ties on 14, and regresses on 12.
The corresponding higher/equal/lower counts are $91/47/14$ against A-Mem, $87/53/12$
against MemFit, and $50/79/23$ against HiGMem (see
Figure~\ref{fig:empirical-accuracy}(b)). Against the strongest aggregate baseline,
HiGMem, H-Mem+ therefore improves 32.9\% of questions and regresses on 15.1\%, while
52.0\% are unchanged. The aggregate gain reflects a favorable balance of paired changes,
rather than uniform improvement on individual questions.

\begin{table}[!t]
\centering
\footnotesize
\setlength{\tabcolsep}{2pt}
\caption{Matched-scope answer quality on LoCoMo10 with DeepSeek-V4.1-Flash-1017.}
\label{tab:current-comparison}
\begin{tabular*}{\columnwidth}{@{\extracolsep{\fill}}lrrrrr@{}}
\toprule
System & Single-hop & Multi-hop & Temporal & \shortstack[r]{Open-\\domain} & Factual F1 \\
\midrule
\multicolumn{6}{l}{\textit{(a) Development conversation: conv-26}} \\
A-Mem~\cite{xu2025amem} & 0.2955 & 0.1173 & 0.2763 & 0.2071 & 0.2458 \\
H-Mem~\cite{sun2026h} & 0.1748 & 0.0792 & 0.1325 & 0.0845 & 0.1366 \\
MemFit~\cite{piehl2026memfit} & 0.2984 & 0.1255 & 0.3502 & 0.1687 & 0.2635 \\
HiGMem~\cite{cao2026higmem} & 0.5617 & 0.3744 & 0.5429 & 0.1683 & 0.4841 \\
\textbf{H-Mem+} & \textbf{0.6471} & \textbf{0.4012} & \textbf{0.6788} & \textbf{0.1766} & \textbf{0.5628} \\
Improvement & $+15.2\%$ & $+7.1\%$ & $+25.0\%$ & $-14.7\%$ & $+16.3\%$ \\
\midrule
\multicolumn{6}{l}{\textit{(b) Five complete conversations (999 questions, 762 factual)}} \\
System & Single-hop & Multi-hop & Temporal & \shortstack[r]{Open-\\domain} & Factual F1 \\
\midrule
A-Mem~\cite{xu2025amem} & 0.3082 & 0.1923 & 0.2709 & 0.1312 & 0.2683 \\
H-Mem~\cite{sun2026h} & 0.1587 & 0.0969 & 0.1111 & 0.0992 & 0.1338 \\
MemFit~\cite{piehl2026memfit} & 0.4102 & 0.1921 & 0.4095 & 0.0940 & 0.3503 \\
\textbf{H-Mem+} & 0.6745 & 0.4425 & 0.6127 & 0.2428 & 0.5926 \\
\midrule
\multicolumn{6}{l}{\textit{(c) All ten conversations (96 Open-domain questions)}} \\
System & Questions & Micro F1 & Macro F1 & BLEU-1 & Failures \\
\midrule
A-Mem~\cite{xu2025amem} & 96 & 0.1646 & 0.1648 & 0.2060 & 0 \\
H-Mem~\cite{sun2026h} & 96 & 0.0909 & 0.0864 & 0.0666 & 0 \\
MemFit~\cite{piehl2026memfit} & 96 & 0.1726 & 0.1692 & 0.1999 & 2 \\
\textbf{H-Mem+} & 96 & 0.3418 & 0.3492 & 0.4314 & 0 \\

\bottomrule
\end{tabular*}
\end{table}

In Table~\ref{tab:current-comparison}, the Open-domain micro mean weights questions, whereas
the macro mean weights equally the nine conversations containing Open-domain questions.
Bold identifies H-Mem+ throughout the matched comparisons.

Across all ten dialogues, ReadAgent attains factual F1 of $0.3899$ with $216/446$
correct refusals, and MemoryBank attains $0.2900$ with $351/446$. On the same questions,
H-Mem+ attains $0.5995$ and $335/446$. MemoryBank achieves more correct refusals than
ReadAgent while recording lower factual overlap. The HiGMem run summary reports
factual F1 of $0.5094$, BLEU-1 of $0.5533$, and $328/446$ correct refusals, with 51
failed question outcomes retained. HiGMem attains higher multi-hop F1 than H-Mem+
($0.4479$ versus $0.4348$).

We report the current A-Mem and MemFit runs on all 999 questions in the five
completed conversations conv-26, conv-30, conv-41, conv-42, and conv-43
(see Table~\ref{tab:current-comparison}(b)). Their 762 factual questions yield F1
$0.2683$ for A-Mem, $0.3503$ for MemFit, $0.1338$ for H-Mem, and $0.5926$
for H-Mem+. All seven MemFit failures are scored as zero. The scope is fixed by
completed conversations rather than performance and is not presented as a random sample
or as a substitute for the ten-dialogue evaluation.

The entire Open-domain category is now available for these four methods: 96 questions
across the nine conversations containing this category; conv-30 has none
(see Table~\ref{tab:current-comparison}(c)). Pooled F1 is $0.1646$ for A-Mem,
$0.1726$ for MemFit, $0.0909$ for H-Mem, and $0.3418$ for H-Mem+.
The two MemFit retrieval truncations remain zero-score failures in the original denominator.
A-Mem exceeds H-Mem+ on conv-44 ($0.1795$ versus $0.1657$).
Micro and conversation-macro averages are reported separately; no partial full-benchmark
score is substituted for a completed run. The rebuilt baseline memories and 16,384-token
completion allowance define a separate run from the historical conv-26 integrations in
panel (a).

We compare H-Mem+ with the preserved A-MEM, MemFit, and HiGMem integrations on conv-26.
H-Mem+ reaches $0.5628$ factual F1 against HiGMem's $0.4841$
(see Table~\ref{tab:current-comparison}(a)). A-Mem, MemFit, and H-Mem obtain factual F1
$0.2458$, $0.2635$, and $0.1366$, respectively. Against HiGMem, H-Mem+ improves
single-hop F1 from $0.5617$ to $0.6471$, multi-hop F1 from $0.3744$ to $0.4012$,
and temporal F1 from $0.5429$ to $0.6788$. Open-domain F1 is $0.1766$, below A-Mem's
$0.2071$, so the advantage does not extend to every category.
HiGMem finalizes all 419 turns and recovers eight of 199 initially truncated answers by
increasing the completion allowance from 2,048 to 4,096. A-MEM and MemFit retain 11 and 13
failed questions, respectively, scored as zero. This comparison evaluates recorded integrations with unequal failure recovery and therefore
does not establish a fully controlled architectural ranking.
A ten-dialogue H-Mem+ aggregate is not substituted for its conv-26 score. The improvement
row uses the strongest other system separately in each category; the overall factual improvement
is $16.3\%$, while open-domain changes by $-14.7\%$ relative to A-Mem. Bold marks H-Mem+,
not an assertion that it wins every category. The current A-Mem and MemFit integrations have now completed all 1,986 questions across ten conversations and passed independent source, metric, and cost audits (see Table~\ref{tab:current-main}). Over 1,540 factual questions, their micro-average F1 scores are 0.2725 and 0.3757, respectively, with BLEU-1 scores of 0.3049 and 0.4150. A-Mem completes all questions successfully; MemFit retains 14 metered length-limit failures as zero scores in the original denominators. Correct refusals are 288/446 and 248/446, respectively.

\subsection{Paired API Token Substudy}
\label{sec:v2-sampled}
\label{sec:higmem-cost}
\label{sec:cost}
\label{sec:efficiency}
To measure token consumption without rebuilding the entire HiGMem benchmark, we analyze
existing responses on conv-26. Within each of the five categories, questions are ordered
by SHA256 of \texttt{seed/category/qid}, with seed zero, and the first ten are selected.
This retrospective sample uses only category and question identifiers, not answers, F1,
or token costs. It is a balanced 50-question substudy of one development conversation,
not a random sample of conversations.

\begin{table}[!t]
\centering
\footnotesize
\setlength{\tabcolsep}{2pt}
\caption{Online API token costs on LoCoMo10.}
\label{tab:sampled-tokens}
\begin{tabular*}{\columnwidth}{@{\extracolsep{\fill}}lrrr@{}}
\toprule
System & API tokens & Tokens/query & API calls \\
\midrule
\multicolumn{4}{l}{\textit{conv-26 (50 questions)}} \\
LoCoMo~\cite{maharana2024locomo} & 1,590,587 & 31,811.7 & 50 \\
ReadAgent~\cite{lee2024readagent} & 331,391 & 6,627.8 & 54 \\
MemoryBank~\cite{zhong2024memorybank} & 303,739 & 6,074.8 & 50 \\
A-Mem~\cite{xu2025amem} & 51,357 & 1,027.1 & 50 \\
H-Mem~\cite{sun2026h}$^{a}$ & 89,957 & 1,799.1 & 50 \\
MemFit~\cite{piehl2026memfit} & 240,431 & 4,808.6 & 100 \\
HiGMem~\cite{cao2026higmem} & 1,240,392 & 24,807.8 & 750 \\
\textbf{H-Mem+} & 61,467 & 1,229.3 & 50 \\
\midrule
\multicolumn{4}{l}{\textit{conv-30/49 (45 questions)}} \\
LoCoMo~\cite{maharana2024locomo} & 1,463,291 & 32,517.6 & 45 \\
ReadAgent~\cite{lee2024readagent} & 260,694 & 5,793.2 & 47 \\
A-Mem~\cite{xu2025amem} & 45,039 & 1,000.9 & 45 \\
H-Mem~\cite{sun2026h}$^{a}$ & 79,730 & 1,771.8 & 45 \\
MemFit~\cite{piehl2026memfit} & 209,709 & 4,660.2 & 88 \\
HiGMem~\cite{cao2026higmem}$^{b}$ & 1,421,379 & 31,586.2 & 681 \\
\textbf{H-Mem+} & 56,351 & 1,252.2 & 45 \\
\bottomrule
\end{tabular*}
\end{table}

Superscript $a$ marks H-Mem's answer-only usage recovered from consecutive cumulative counters,
and $b$ marks HiGMem's incomplete-input condition (see Table~\ref{tab:sampled-tokens}).
LoCoMo, ReadAgent, and MemoryBank use a 4,096-token completion cap.
The same fixed question IDs also admit complete online costs for the newly audited
integrations. On the 50-question conv-26 sample, LoCoMo, ReadAgent and MemoryBank use
1,590,587, 331,391 and 303,739 API tokens in 50, 54 and 50 calls, respectively. On the
fixed 45 questions from conv-30/49, LoCoMo uses 1,463,291 tokens in 45 calls and
ReadAgent uses 260,694 tokens in 47 calls. MemoryBank has no complete matched
45-question sample, so no value is imputed. These columns sum the original online
responses and any failures, exclude construction and local computation, and do not
require new API calls. An old ReadAgent conv-50 HTTP402 request has unknown usage;
it is outside both fixed samples, remains disclosed as a ledger gap, and its separate
33,616-token reservation is not reported as measured cost.

We sum API-reported input and output tokens for every retrieval and answer call attached
to a selected question, including recorded failed attempts and recovery. Reasoning and
cache tokens are subsets and are not added again. H-Mem+ uses 61,467 tokens in 50 calls;
HiGMem uses 1,240,392 tokens in 750 calls. The ratio of summed costs gives a $95.04\%$
reduction. A 5,000-draw paired bootstrap within categories gives a conditional 95\% interval
of $94.50\%$--$95.51\%$. It does not capture uncertainty across conversations or new runs.
The same 40 factual questions have F1 $0.4695$ for H-Mem+ and $0.3911$ for HiGMem;
the ten adversarial questions produce eight and six correct refusals, respectively.
Thus the observed online token reduction is not accompanied by lower factual F1 in this sample.

\begin{figure*}[!t]
\centering
\includegraphics[width=\textwidth]{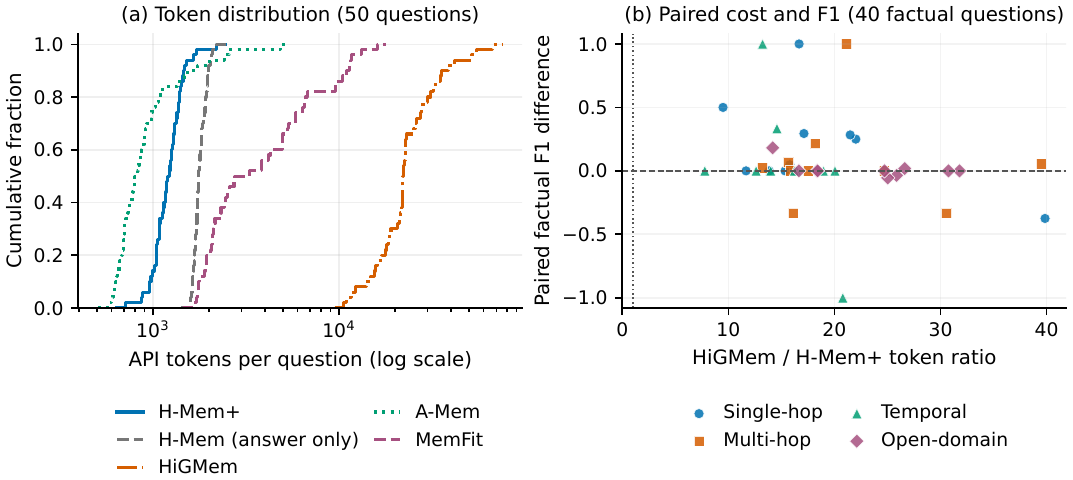}
\caption{Per-question API cost and answer quality on the fixed conv-26 sample.
(a) Cumulative online token distributions for 50 questions. H-Mem uses historical answer-only
counters. The other curves count retrieval and answering. (b) HiGMem-to-H-Mem+ token ratios and
paired factual F1 differences (H-Mem+ minus HiGMem) on the 40 factual questions. Ratios above
one indicate lower token use for H-Mem+. H-Mem+ uses fewer tokens on all 40 questions, with higher
F1 on 15, equal F1 on 19, and lower F1 on six. The costs exclude memory construction and local computation.}
\label{fig:empirical-tokens}
\end{figure*}
We examine the distributions of online token costs and paired F1 differences to assess
whether aggregate savings are consistent across questions. H-Mem+ has a median online cost of 1,212 tokens per question (range 711--2,198),
compared with 22,102 for HiGMem (range 10,631--67,175). The corresponding medians are
811 for A-Mem, 3,032.5 for MemFit, and 1,765 for the recorded H-Mem answer counters
(see Figure~\ref{fig:empirical-tokens}). These distributions confirm that the lower
summed cost relative to HiGMem is accompanied by a lower median, while A-Mem remains
less costly by both measures. On the 40 factual questions, the mean F1 difference
between H-Mem+ and HiGMem is $0.0784$ ($0.4695$ versus $0.3911$); this aggregate
advantage does not imply that every paired answer improves. For the same question
identifiers, we recover H-Mem's answer tokens by differencing consecutive cumulative usage
snapshots within each conversation (see Table~\ref{tab:sampled-tokens}). Consecutive question IDs,
one added call per question, nonnegative increments, and constant memory counters are checked.
The audited A-Mem and MemFit runs provide costs for all 50 questions. They use 51,357 and 240,431 online tokens, respectively: H-Mem+ uses
$74.43\%$ fewer tokens than MemFit but $19.68\%$ more than A-Mem on this sample.
Thus the cost advantage is comparator-specific. A-Mem and MemFit use a 16,384-token
completion allowance after known, metered truncations in technical validation, whereas
H-Mem+ and HiGMem use 4,096 for their readers. These are measured integration costs,
not an equal-output-cap causal comparison. The separately audited memory-stage
costs are 1,730,356 and 113,401 tokens, respectively, and are excluded from the online table.

The fixed 50-question comparison covers online costs and excludes local computation.
Selected query-stage usage is complete. H-Mem+ reuses historical memory with an incomplete
construction ledger, and the HiGMem conv-26 ledger lacks usage for its first 45 turns and
an interrupted legacy call. The paired comparison therefore does not establish end-to-end savings.
Different reader-cap recovery and construction histories remain
qualifications. All 1,986 H-Mem+ answer calls separately record 2,323,938 tokens, but that
number is not an end-to-end cost and is not compared with HiGMem's construction-plus-query
cost. The HiGMem ten-dialogue run summary records $46{,}747$ calls and $88{,}980{,}962$
observed tokens, including failed and recovered attempts. Eighteen of those calls did not
return prompt-token counts, so the figure is a lower bound on its cost rather than a closed total.

\subsection{Reader Sensitivity and Controlled Ablations}
\label{sec:final-ablations}
\begin{table*}[!t]
\centering
\footnotesize
\setlength{\tabcolsep}{5pt}
\caption{Controlled H-Mem+ ablations on 180 questions from nine non-development conversations.}
\label{tab:controlled-ablation}
\begin{tabular}{lrrrrrr}
\toprule
Configuration & Evidence budget & Factual F1 & $\Delta$ F1 & BLEU-1 & Correct refusal & Tokens/question \\
\midrule
H-Mem+ (control) & 2,048 & 54.80 & +0.00 & 60.87 & 25/36 & 1,235.4 \\
Without raw candidate bypass & 2,048 & 49.90 & -4.90 & 57.22 & 25/36 & 1,502.1 \\
Without CE (weighted RRF) & 2,048 & 43.55 & -11.24 & 49.72 & 25/36 & 1,362.4 \\
Without CE (candidate-index order) & 2,048 & 12.41 & -42.39 & 17.49 & 33/36 & 1,300.5 \\
Smaller evidence budget & 1,024 & 55.04 & +0.24 & 61.26 & 25/36 & 1,221.4 \\
Larger evidence budget & 4,096 & 54.80 & +0.00 & 60.87 & 25/36 & 1,235.4 \\
\bottomrule
\end{tabular}
\end{table*}

With the evidence held fixed across all 1,986 questions, factual F1 is $0.5995$
with DeepSeek and $0.4417$ with Qwen. The difference shows that answer quality
depends on the reader even when both models receive the same evidence.

For Table~\ref{tab:controlled-ablation}, factual F1, BLEU-1, and paired $\Delta$F1 use
percentage points; adversarial questions are excluded from factual scores and their correct
refusals are reported as counts. Evidence budgets use BERT wordpieces, distinct from API tokens.
Tokens per question include prompt and completion usage, with reused reference usage counted
in each arm's effective cost, and exclude memory construction and local retrieval.
The no-cross-encoder replacement uses weighted reciprocal-rank fusion on the same candidate
pool, with $k=60$, summary dense/lexical weights $0.25/0.25$, and raw lexical weight $0.5$.
Candidate-index order is retained separately as a degradation diagnostic.
Identical prompts reuse saved control responses. In particular, the 4,096-wordpiece setting
changes no prompt relative to the 2,048-wordpiece control and requires zero new API calls;
the larger evidence budget is non-binding on this sample and is not an independent replicate.

We additionally evaluate single-factor ablations on a fixed 180-question non-development sample (see Table~\ref{tab:controlled-ablation}). Removing the raw candidate bypass reduces factual F1 by 4.90 percentage points (paired dialogue-cluster bootstrap 95\% interval: $[-8.64,-0.63]$); replacing cross-encoder reranking with the fixed weighted-RRF rule changes F1 by $-11.24$ points ($[-17.23,-5.78]$). The candidate-index-order diagnostic yields F1 of 12.41, but truncates away all raw-window candidates and cannot isolate the contribution of the cross-encoder. Reducing the evidence budget to 1,024 wordpieces changes F1 by $+0.24$ points ($[-0.17,+0.88]$), whereas the larger-budget row follows the exact-reuse policy described above. These conditional intervals use nine dialogue clusters and 5,000 bootstrap draws; a single response per distinct prompt does not establish variation across repeated generations. Blank result cells indicate pending complete audits, not zero scores; no partial run or projected result is substituted for a completed evaluation.

For the same 45-question sample, A-Mem uses 45,039 online tokens in 45 calls and MemFit uses 209,709 tokens in 88 calls, including metered failures. H-Mem+ therefore uses more online tokens than A-Mem on this sample; its cost advantage is not uniform across baselines. On the 45 paired questions from conv-30 and conv-49 (see Table~\ref{tab:sampled-tokens}), H-Mem+ uses 56,351 online API tokens versus 1,421,379 for HiGMem, a 96.04\% reduction in retrieval-plus-answer costs. On the 35 factual questions, micro-average F1 is 0.5744 and 0.5548, respectively. HiGMem omits four explicitly recorded turns after truncated memory-generation responses: turns 171, 181, and 353 of conv-30 and turn 380 of conv-49. These responses are not retried; their costs remain included in the construction ledger. Thus, this result is conditional on incomplete HiGMem memory input. Its observed construction cost is 5,944,634 tokens, including imported and failed calls. Construction and local computation are excluded from the online comparison; the sample does not establish full-benchmark HiGMem accuracy or end-to-end cost savings.



\subsection{Discussion}
\label{sec:discussion}
\label{sec:disc-evidence}
H-Mem+ separates memory organization from evidence delivery. A hierarchy can organize
summaries while the reader selects source facts through broad candidates and local ranking.
HiGMem is also hierarchical; its performance does not support a general claim that hierarchy
cannot help accuracy. Our comparison instead demonstrates a different allocation of online
API work: local candidate generation and reranking replace generative retrieval calls, and
the preserved 50-question sample uses substantially fewer API tokens at higher factual F1.
Whether that trade-off transfers across conversations remains an empirical question.

\noindent\label{sec:disc-limits}The theoretical statements have different scopes. Ancestor-index
entropy is not semantic-content entropy. A deterministic summary chain obeys data processing,
but the original-record bypass avoids making it the only information path. A beam-routing
score certificate is not a set-recall or answer-F1 guarantee. The mean-aggregation stability
result is conditional and is not established for the implementation's LLM-generated summaries.
Likewise, the depth optimum is for a routing surrogate, not an optimum for H-Mem+'s API bill.
These distinctions preserve the useful design consequences without assigning unproved
guarantees to the implementation. Multi-source coverage and budget packing, rather than
best-score approximation alone, are the quantities most directly aligned with answering.

\section{Conclusion}\label{sec:conclusion}
H-Mem+ combines a hierarchical summary index with a source-preserving evidence path,
broad candidate generation, local reranking, and bounded context packing. The analysis
separates conditional properties of abstraction and routing from candidate availability,
packing omissions, and metered API costs. On LoCoMo10, the final method achieves factual
F1 $0.5995$ and correct refusal $335/446$. On a fixed retrospective 50-question sample from
conv-26, retrieval-plus-answer API tokens fall by $95.04\%$ relative to our HiGMem integration,
without lower factual F1 on that sample. This is not a full-benchmark end-to-end cost claim.
Future work should measure construction cost with complete ledgers, repeat paired cost
comparisons outside the development conversation, and evaluate the frozen method across
backbones and independent datasets. Rate--distortion optimality and online regret remain
open, rather than being consequences of the conditional hierarchy bounds.



\end{document}